\documentclass[conference]{IEEEtran}
\IEEEoverridecommandlockouts
\usepackage{cite}
\usepackage{amsmath,amssymb,amsfonts}
\usepackage{algorithmic}
\usepackage{graphicx}
\usepackage{textcomp}
\usepackage{xcolor}
\def\BibTeX{{\rm B\kern-.05em{\sc i\kern-.025em b}\kern-.08em
    T\kern-.1667em\lower.7ex\hbox{E}\kern-.125emX}}
\begin{document}

\title{Constraint-based Formation of Drone Swarms}

\author{\IEEEauthorblockN{Xijing Liu\IEEEauthorrefmark{1},
Kevin Lam\IEEEauthorrefmark{2}, Balsam Alkouz\IEEEauthorrefmark{3},
Babar Shahzaad\IEEEauthorrefmark{4}, and
Athman Bouguettaya\IEEEauthorrefmark{5}}
\IEEEauthorblockA{School of Computer Science,
The University of Sydney\\
Australia\\
Email: \IEEEauthorrefmark{1}xliu4593@uni.sydney.edu.au,
\IEEEauthorrefmark{2}lamkevin042@gmail.com,
\IEEEauthorrefmark{3}balsam.alkouz@sydney.edu.au,
\IEEEauthorrefmark{4}babar.shahzaad@sydney.edu.au,\\
\IEEEauthorrefmark{5}athman.bouguettaya@sydney.edu.au
}}






\maketitle

\begin{abstract}
Drone swarms are required for the simultaneous delivery of multiple packages. We demonstrate a multi-stop drone swarm-based delivery in a smart city. We leverage formation flying to conserve energy and increase the flight range of a drone swarm. An adaptive formation is presented in which a swarm adjusts to extrinsic constraints and changes the formation pattern in-flight. We utilize the existing building rooftops in a city and build a line-of-sight skyway network to safely operate the swarms. We use a heuristic-based A* algorithm to route a drone swarm in a skyway network.  
\end{abstract}


\section{Introduction}
Drone swarms are used in a wide range of applications where a single drone would be inadequate \cite{cardona2019robot}. A drone swarm can collectively execute complex tasks with more efficiency and reduced cost \cite{alkouz2020formation}. They are typically used in search and rescue, sky shows, surveillance, and delivery \cite{shahzaad2021robust} \cite{waibel2017drone}. The focus of this paper is on the use of drone swarms to deliver multiple packages simultaneously to the same destination. A delivery swarm is utilised to circumvent the limitations imposed by current flight regulations, which only permit the operation of small drones in cities\footnote{https://www.faa.gov/uas/advanced\_operations/package\_delivery\_drone}. Therefore, the delivery of multiple/heavier packages simultaneously necessitates the use of drone swarms, as larger drones cannot be deployed. A swarm in delivery is defined as a group of drones that move together from a source to a destination to achieve a common objective, i.e., the delivery of packages \cite{alkouz2021reinforcement}. A drone swarm delivery operation is constrained by a number of factors \cite{alkouz2022service}. For instance, the drones' payload and battery limitations are regarded as intrinsic constraints. The weather conditions and availability of recharging pads at recharging stations are extrinsic constraints \cite{shahzaad2019composing}. Furthermore, each drone in a swarm may carry a different payload resulting in a variation in battery consumption rates. The optimal routing of a drone swarm in a skyway network must cater for all the aforementioned constraints.\looseness=-1

\begin{figure}[htb!]
\centering
\includegraphics[width=0.9\linewidth]{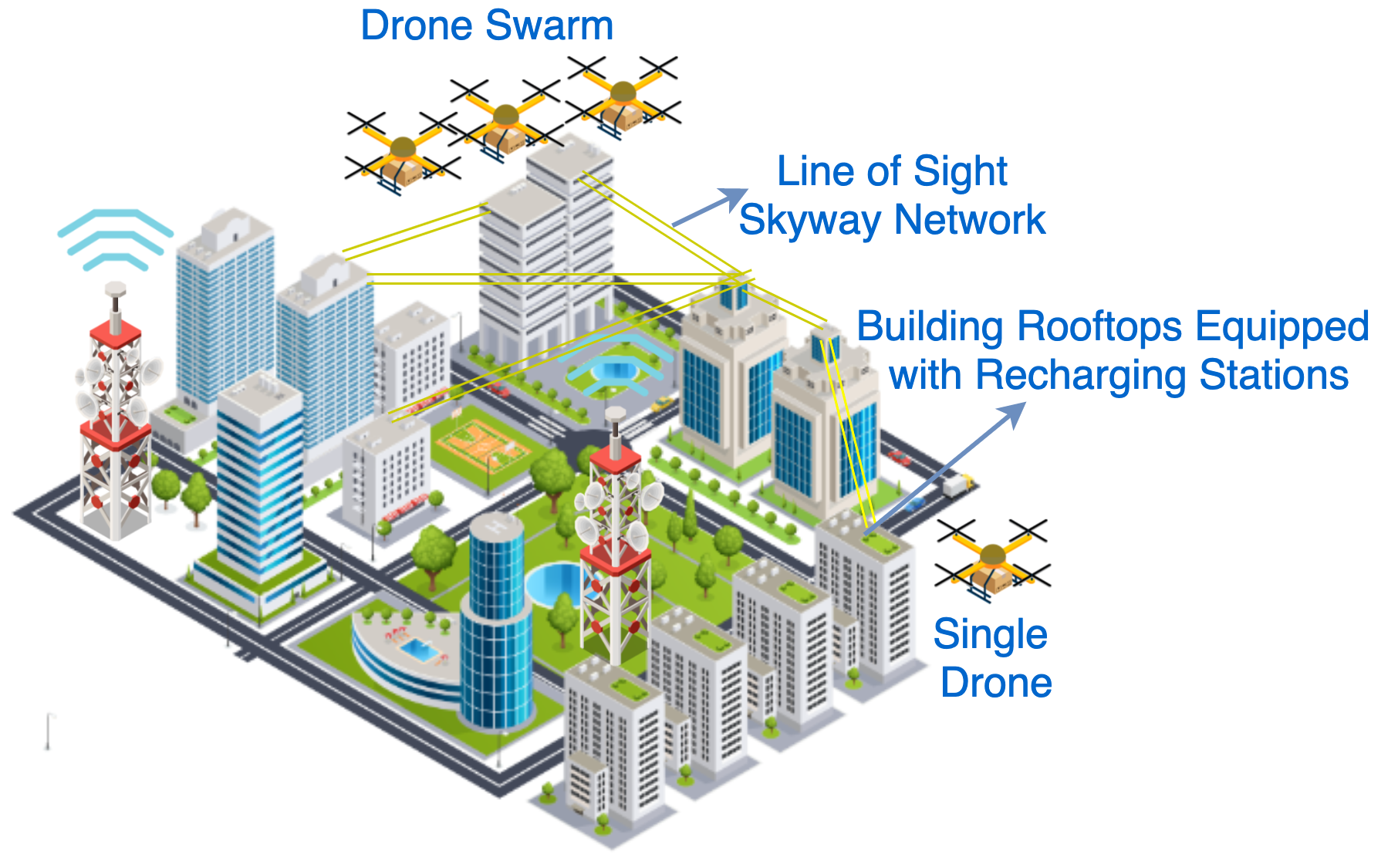}
\caption{A Skyway Network Infrastructure}
\label{skyway}
\end{figure}

Fig. \ref{skyway} depicts the typical operating environment of drown swarms in a skyway network. Building rooftops represent nodes in the skyway network \cite{10.1145/3460418.3479289}. They can also act as recharging pads. Each node can be a source, destination, or intermediate recharging station. Skyway segments are line-of-sight paths that directly connect two nodes within each other's sight \cite{jermaine2021demo}. We present a novel definition of a drone swarm service as as packages delivery between two nodes along a skyway segment. We assume that a swarm may take different \textit{formations} to optimize the service delivery.

Formation flying is inspired by the natural behaviour of birds and insects \cite{mirzaeinia2019energy}. Various aspects of flight formation such as aerodynamics, control, and energy conservation, have been studied based on migrating birds \cite{cutts1994energy}. According to several studies, migrating geese save energy by flying in a V-shaped formation \cite{cutts1994energy}. In this regard, the aerodynamic effect of formation flying in drone swarms has been investigated. Drone swarms flying in formation save energy due to drag forces and translational lift provided by upwash/downwash forces from neighbouring drones \cite{mirzaeinia2019energy}. An ideal formation conserves the most energy under certain wind conditions, hence improving the swarm flying range. We \textit{demonstrate an adaptive swarm} that alters its formation based on the wind conditions \footnote{Demo: https://youtu.be/NYv31r6zC\_Y}. Drones in a formation consume different amounts of energy due to two factors: the carried payload and the position in the formation \cite{alkouz2020formation}. When planning the route, we take into account the different energy consumption rates of drones within a swarm.\looseness=-1



\section{Demonstration System}
\subsection{Hardware Components and Lab Setup}
\begin{figure}[ht]
\centering
\includegraphics[width=0.8\linewidth]{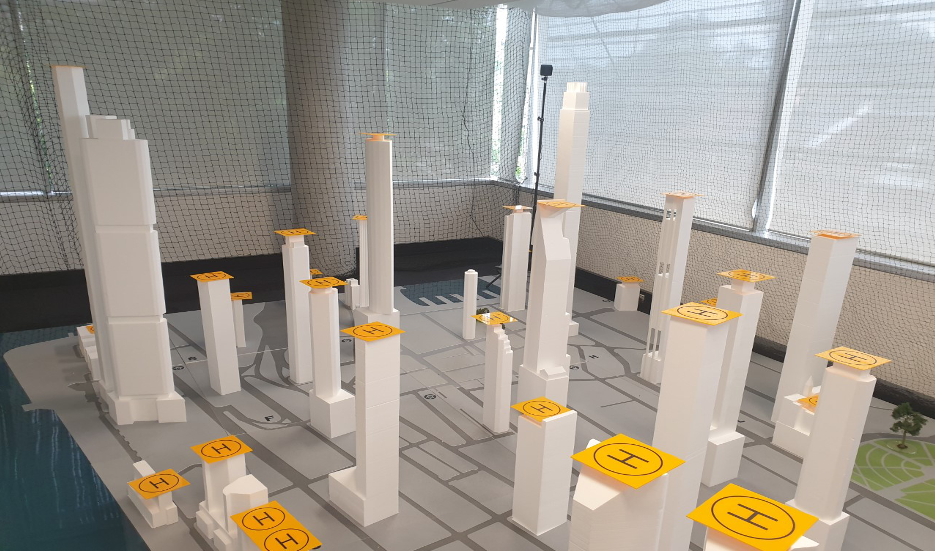}
\caption{3D Model of City CBD}
\label{3d}
\end{figure}

\begin{figure}[ht]
\centering
\includegraphics[width=0.8\linewidth]{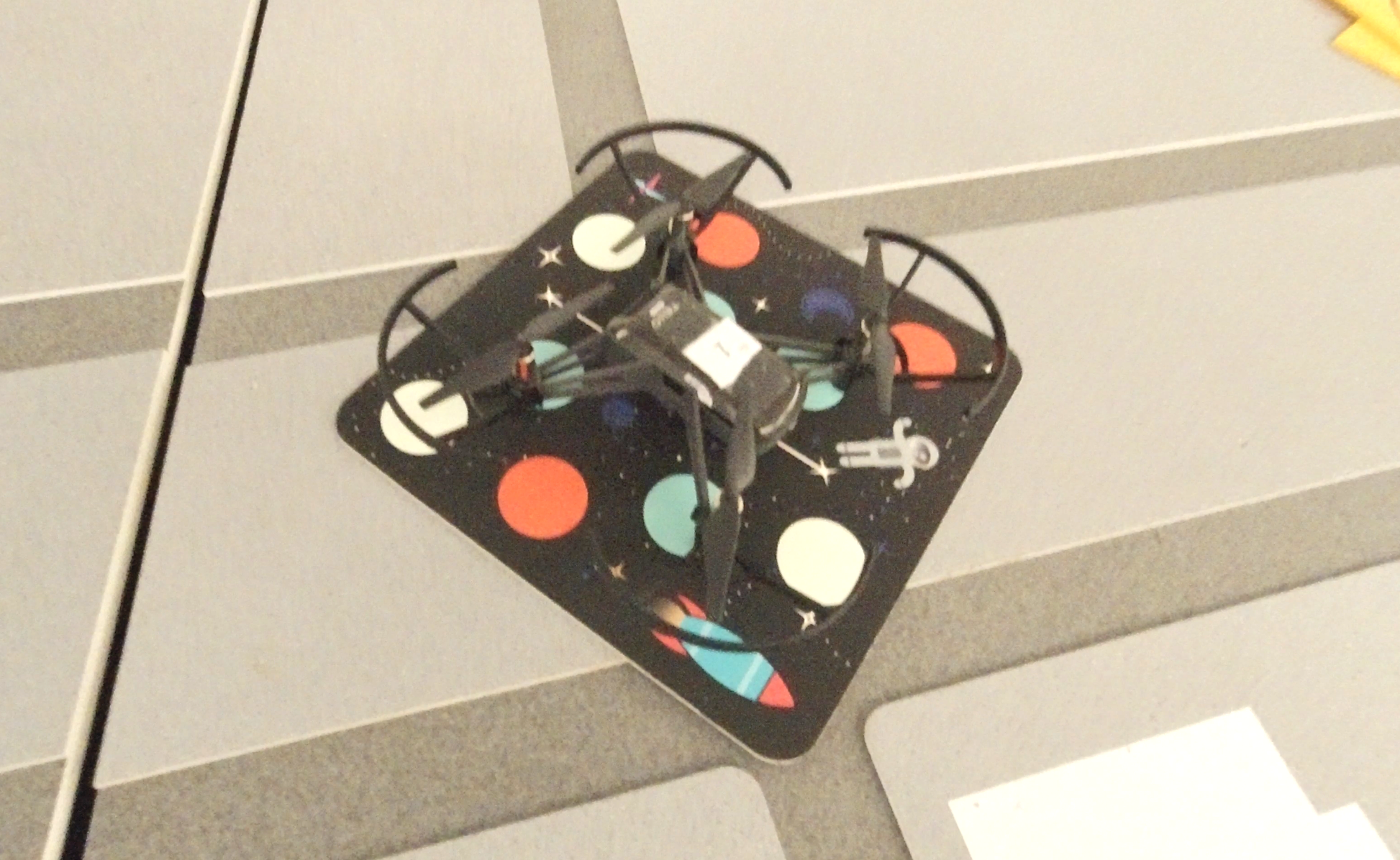}
\caption{Tello Drone on a Mission Pad}
\label{Drone}
\end{figure}

We present an indoor testbed environment for the deployment of swarm-based drone deliveries. Outdoor testing of drone swarm is laden with danger. In addition, aviation regulations restrict the usage of drones in populated regions \cite{jones2017international}. Therefore, an indoor drone testbed has been built as part of the demonstration (Fig. \ref{3d}). The setup consists of 3D printed buildings of a City CBD. Each rooftop is assumed to be a recharging node or a delivery point in a skyway network. We use five DJI Edu Tello drones, which are safe to operate indoors due to their small size (Fig. \ref{Drone}).
The Tello drones are equipped with a Vision Positioning System (VPS). A VPS enables a drone to map the ground while in flight, allowing it to know its position in relation to the ground. To locate a drone's position, the VPS employs a combination of visual and ultrasonic waves that are bounced off the ground. For a swarm setup, a router/access point is required to communicate with all of the drones when flying in a swarm. The swarm commands are issued through a Python code that goes through the router to each of the Tello drones. We first reconfigure each drone in the swarm individually and set them to station mode. We use the Packet Sender\footnote{https://packetsender.com/} software to connect to the Tello's WiFi and put the Tellos into SDK mode to send swarming commands.

\subsection{Software Components}
Four major components are involved in programming a swarm to fly in formation within a skyway network. These components are (1) line-of-sight skyway network, (2) optimal route planning algorithm, (3) formation flying mechanism, and (4) real-time swarm monitoring interface. Each of these components is described below.

\subsubsection{Line-of-Sight Skyway Network}
\begin{figure}[ht]
\centering
\includegraphics[width=0.9\linewidth]{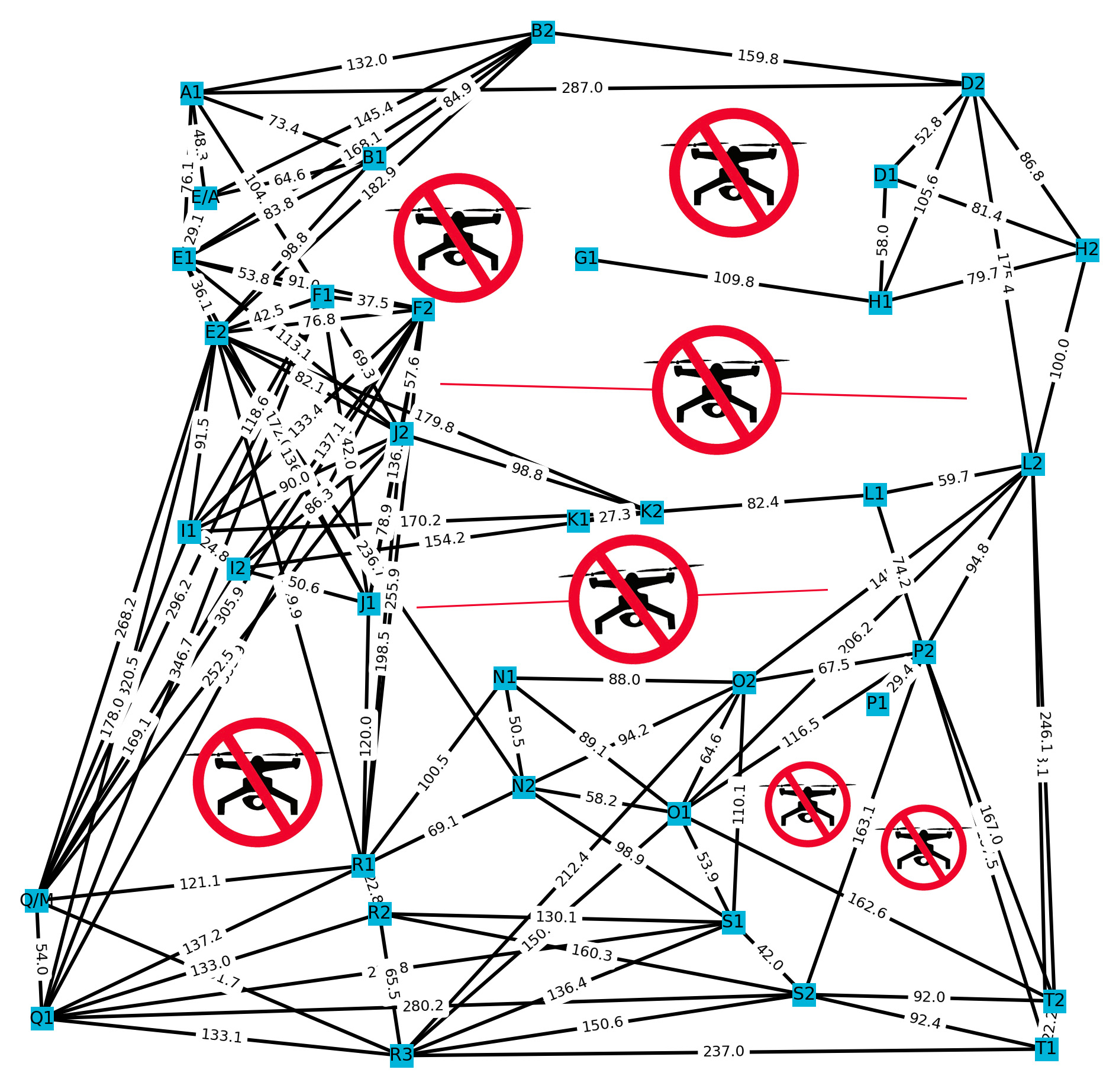}
\caption{Line of Sight Paths in a Skyway Network with No-Flight Zones}
\label{network}
\end{figure}

A skyway network, as described earlier, consists of connected building rooftops through skyway segments. The purpose of this component is to compute the line-of-sight paths between the rooftops. If a tall building ($X$) blocks two other shorter buildings ($A$ and $B$), no path could be established between $A$ and $B$. In addition, if a no-flight zone intersects the buildings $A$ and $B$, no path could be established either. According to flight regulations, drones are not permitted to fly in no-flight zones for security or public safety concerns\footnote{https://www.casa.gov.au/drones/drone-rules/flying-near-emergencies-and-public-spaces}. A no-flight zone could be located near airports, national parks, marine and wildlife habitats, etc. As indicated in Fig. \ref{network}, we define seven no-flight zones in our designed skyway network.

The line-of-sight method examines intersections with the other buildings in the region based on the X,Y,Z coordinates of the buildings. The highest point of the building, where the swarm would generally hover or land to recharge, is denoted as the Z coordinate. To check whether a node is directly reachable from the current node or not, the heuristic-based algorithm runs several steps. First, the algorithm creates a rectangular polygon for the given swarm width that symbolises the path the swarm will follow between the current node and the target node. Second, it identifies all nodes in the area that intersect with this polygon. This step solely considers the node's XY plane. The first two steps are designed to reduce the processing time by only considering potential nodes. Third, a right angle triangular polygon is created using the heights (Z coordinates) of the current node and the target node. Considering all the potential nodes from the second step, we find if these nodes block the path between the current node and the target node. If a node lies above the triangular polygon computed in step three, then this node blocks the path. Therefore, no line-of-sight path could be established between the current and target nodes. Otherwise, if the node lies within or below the polygon, a line-of-sight path is established. Fig. \ref{network} shows the line-of-sight skyway network computed from the indoor testbed building rooftops.\looseness=-1

\subsubsection{Optimal Route Planning Algorithm}

\begin{figure}[htp!]
\centering
\includegraphics[width=0.9\linewidth]{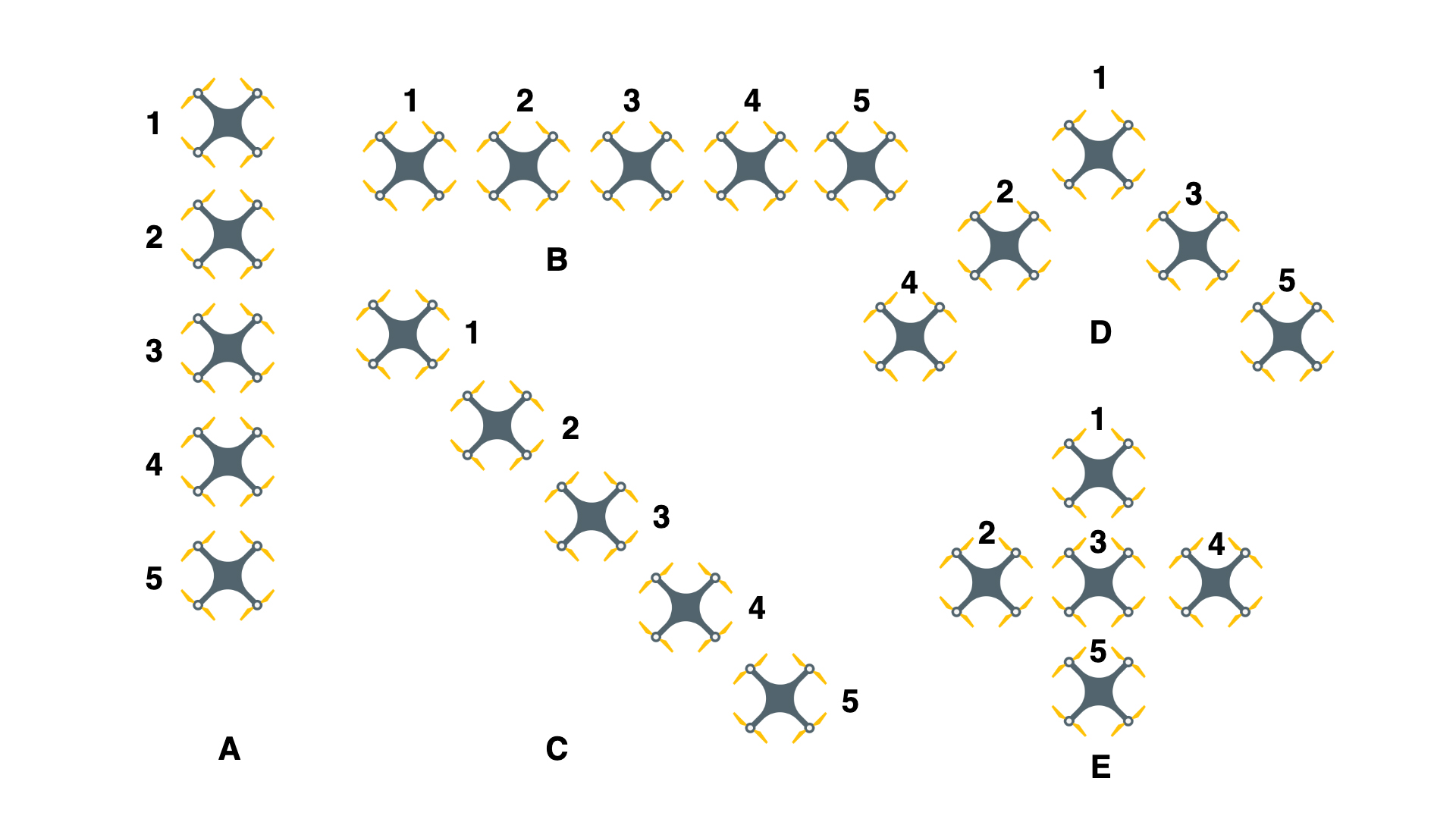}
\caption{Different Swarm Flight Formations}
\label{formations}
\end{figure}

An effective route planning ensures that a swarm traverses the network in the least amount of time between the source and the destination. For simplicity, we assume that the environment is deterministic, i.e., the availability of recharging pads and the wind conditions are known. We implement a heuristic-based A* algorithm to optimally compose the path between the source and destination. The algorithm also selects the best formation at a certain time based on the wind conditions and the results reported in \cite{alkouz2020formation}. We assume that a swarm may be shaped in five different formations, i.e., column, front, echelon, vee, and diamond (Fig.\ref{formations}). This module provides an optimal path, which includes recharging nodes, flown over nodes, and formation changes at certain time intervals.

\subsubsection{Formation Flying Mechanism}
We implement autonomous formation flights using Tello drones with the aid of mission pads (Fig. \ref{Drone}). The drones locate themselves using the VPS system that reads the mission pads distributed in the environment. Each mission pad has a unique pattern and hence a unique coordinate. As the swarm reads the mission pads, it directs its rotation angle towards the next node. The swarm then flies in a direct line-of-sight path. When a formation change instruction is received, relevant drones in the swarm adjust their positions in relation to other drones in order to change the formation.

\subsubsection{Real-Time Swarm Monitoring Interface}

\begin{figure}[ht]
\centering
\includegraphics[width=0.9\linewidth]{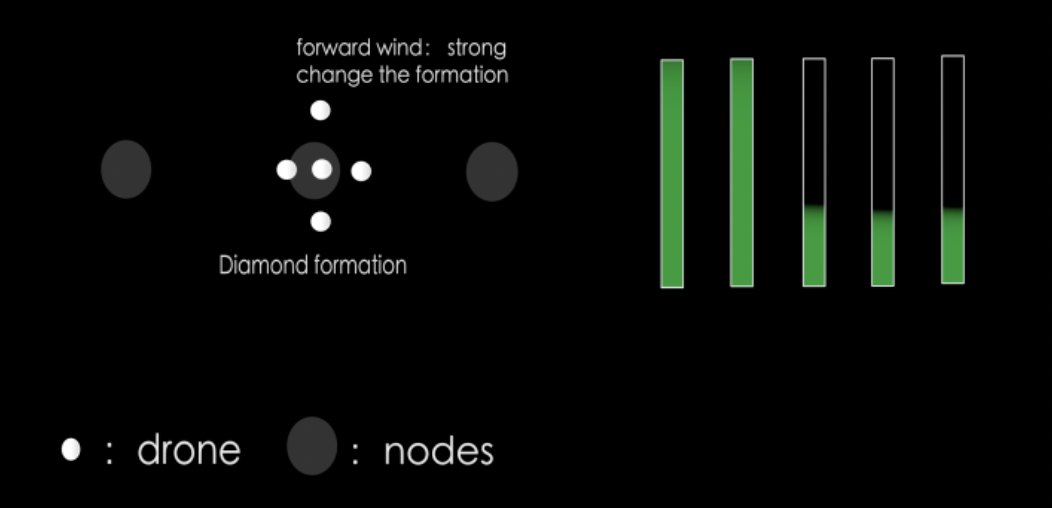}
\caption{Real-Time Swarm Monitoring GUI}
\label{GUI}
\end{figure}

We create a simple user interface to track the swarm during its delivery operation. Once a path is determined using the optimal route planning algorithm, the interface displays the nodes including the recharging nodes. The battery status of each drone is indicated to track each drone's recharging requirements in the swarm. The position and formation changes are reported in real-time. Fig. \ref{GUI} shows the GUI during the drone swarm flight.


\section{Conclusion}
We present a prototype that enables swarm-based drone delivery in a congested city setting. We construct a line-of-sight skyway network that enables a safe drone swarm-based delivery operation. We take into consideration the no-flight zones that could be specified by governments. A heuristic-based A* path composition algorithm is used to compute the optimal path from source to destination in terms of delivery time. A swarm utilizes landmarks to locate and position itself in the indoor testbed environment built of a City CBD. We demonstrate how a drone swarm adjusts its formation in response to changing wind conditions for reducing energy consumption and increasing flight time. Finally, a user interface is designed to track the swarm during its delivery operation. In future, we plan to incorporate obstacles avoidance algorithms for the effective provisioning of drone delivery services.



\bibliographystyle{IEEEtran}  
\bibliography{percom} 

\end{document}